\documentclass{article}

\usepackage[main, final, nonatbib]{neurips_2026}
\usepackage[numbers]{natbib}

\usepackage[utf8]{inputenc} %
\usepackage[T1]{fontenc}    %
\usepackage{hyperref}       %
\usepackage{url}            %
\usepackage{booktabs}       %
\usepackage{amsfonts}       %
\usepackage{nicefrac}       %
\usepackage{microtype}      %
\usepackage{xcolor}         %

\usepackage{amsmath}
\usepackage{amssymb}
\usepackage{mathtools}
\usepackage{amsthm}

\usepackage{algorithm}
\usepackage{algpseudocode}

\usepackage{hyperref}
\usepackage{url}
\usepackage{graphicx}
\usepackage{bm}
\usepackage{multirow}
\usepackage{tikz}
\usepackage{booktabs}
\usepackage{colortbl}
\usepackage{xcolor}
\usepackage[whole]{bxcjkjatype}
\usepackage[framemethod=TikZ]{mdframed}
\usepackage{wrapfig}
\usepackage{cleveref}
\Crefname{section}{\S}{\S\S}

\title{Reconsidering Positional Supervision in \\Masked Diffusion Language Model Training}

\author{%
  Mengyu Ye\thanks{Corresponding to: \nolinkurl{<ye.mengyu.s1@dc.tohoku.ac.jp>}; \nolinkurl{<is-failab-research@grp.tohoku.ac.jp>}} \\
  Tohoku University \\
  \And
  Keito Kudo \\
  Tohoku University \& RIKEN \\
  \AND
  Ryosuke Takahashi \\
  Tohoku University \& RIKEN \\
  \And
  Jun Suzuki \\
  Tohoku University \& RIKEN \& NII LLMC. \\
}

\begin{document}

\maketitle

\begin{abstract}
Masked diffusion language models (MDLMs) generate text by unmasking tokens in parallel and have recently emerged as alternatives to autoregressive language models. 
They can be viewed as parallel decoders trained with a position-wise cross-entropy (CE) loss, the same setup as non-autoregressive translation (NAT). 
In NAT, CE-trained parallel decoders have been argued to be sensitive to small positional shifts, since CE penalizes them harshly.
We ask whether CE-trained MDLMs are similarly sensitive to such shifts under iterative decoding. To probe this, we apply a controlled intervention that introduces them during decoding. On LLaDA-8B-Instruct with Arena-Hard, displacing as little as 1\% of generated tokens by one position substantially reduces win rates against the unintervened model, showing that MDLMs are sensitive to such small shifts under iterative parallel decoding. 
Motivated by this, we adapt connectionist temporal classification (CTC), an alignment-flexible objective known to mitigate it there, to MDLM supervised fine-tuning. 
By relaxing the strict position-wise match that CE imposes, CTC gives the loss room to absorb small positional shifts; concretely, we modified CTC objective to use a special \textsc{\textless slack\textgreater} token that absorbs positional uncertainty between target tokens and output positions, and a updated collapse map that preserves target surface forms.
Across four open-ended generation benchmarks, the resulting model consistently improves over both the original model and a matched cross-entropy-trained baseline, with statistically significant gains on all four.
These results identify training-side alignment flexibility as a useful design dimension for MDLM SFT, complementary to the inference-time approaches explored in prior work\footnote{Code: \url{https://github.com/muyo8692/mdlm-ctc}}.
\end{abstract}

\section{Introduction}

Diffusion language models have recently emerged as promising alternatives for text generation~\cite{nie2025llada,zhu2025llada1p5, song2025seeddiffusion, labs2025mercury, bie2025llada20}. Unlike autoregressive (AR) language models~\cite{vaswani2017attention, brown2020gpt3} that generate text strictly left-to-right, diffusion language models decode multiple tokens in parallel. In particular, Masked Diffusion Language Models (MDLMs) have closed much of the perplexity gap to AR models at small scales~\cite{lou2024sedd, ou2025your, sahoo2024simple, shi2024simplified, arriola2025interpolating}, and at larger scales~\cite{nie2025llada, ye2025dream, zhu2025llada1p5, song2025seeddiffusion, bie2025llada20} match strong AR baselines on challenging math and reasoning tasks.

Standard MDLM training corrupts gold text by randomly masking some tokens, leaves the remaining gold tokens as conditioning context, and supervises the model with a cross-entropy (CE) loss to predict each masked position from the partially masked sequence. This choice simplifies and stabilizes training~\cite{sahoo2024simple}, but it builds positional rigidity into the objective at multiple points: each prediction is aligned to its target at a fixed output position, and the conditioning context surrounding each masked token is assumed to be gold text at gold positions. CE has no mechanism to absorb small displacements at either point. This combination of parallel decoding and position-wise CE is also the defining setup of non-autoregressive translation (NAT)~\cite{gu2018nat}, where CE-trained parallel decoders have been argued to be sensitive to small positional shifts~\cite{ghazvininejad2020axe,du2021order,du2022ngram}: CE penalizes small positional shifts. Whether MDLMs, which share the same parallel-prediction structure and position-wise CE objective, are similarly sensitive to small positional shifts under iterative decoding has not been examined.

We test this directly under open-ended generation, the setting aligned to NAT's task form, where positional shifts have the most room to accumulate over many denoising steps. We apply a minimal positional disturbance during decoding: we swap an unmasked token with an adjacent masked position, so token identities are preserved and only a one-position shift is introduced at the time of intervention. This controlled intervention is diagnostic: it isolates sensitivity to small positional shifts under irreversible decoding rather than modeling all naturally occurring errors. On LLaDA-8B-Instruct~\cite{nie2025llada} with Arena-Hard~\cite{arenahard2024}, shifting as little as 1\% of generated tokens by a single position drops the win rate against the unintervened model well below 50\%. 
This suggests that MDLMs are sensitive to small shifts under iterative parallel decoding.

Motivated by this finding, we adapt a modified form of connectionist temporal classification (CTC)~\cite{graves2006ctc}, an alignment-flexible training objective, to the MDLM SFT setting. CTC has already been applied to NAT~\cite{libovicky2018ctcnat, saharia2020ctcnat} and demonstrably mitigates this sensitivity there. The flexibility comes from a blank symbol: CTC lets the model emit blanks alongside content tokens, and treats output sequences that differ only in blank placement as equivalent. A content token assigned to a nearby position is then absorbed by an adjacent blank without penalty. We instantiate the blank as a special \textsc{\textless slack\textgreater} token and add the CTC objective as an auxiliary loss alongside the standard CE loss during supervised fine-tuning (SFT); \textsc{\textless slack\textgreater} tokens are stripped from outputs after decoding. To preserve target surface forms, we also modify standard CTC's collapse rule, which merges consecutive repeated content tokens and would otherwise distort outputs (collapsing ``**'' into ``*'' or ``100'' into ``10''), so it removes only blank symbols.

Applied to LLaDA-8B-Instruct, our method improves over both the original model and a matched CE baseline on all four tested open-ended generation benchmarks (Arena-Hard, MT-Bench, WildBench and Creative Writing Bench), with gains that are statistically significant on all four. 
Our contribution is to bridge NAT and MDLM in the SFT setting: motivated by their shared parallel-CE structure, we empirically show that training-side alignment flexibility known to help NAT also helps MDLMs. 
In contrast to prior work on MDLM decoding failures, which has mostly operate at inference time~\cite{wu2025dreamon, wang2025remasking}, these results identify alignment flexibility as a complementary training-side design dimension for MDLM SFT.

\section{Masked Diffusion Language Models}
\label{sec:background}
MDLMs define a probability distribution over sequences through a forward noising process and a learned reverse denoising process. 
Let \( x_0 = (x_0^1,\dots,x_0^L) \) be a token sequence, whose length is $L$, and contain no masks.
The forward process independently masks each token with a probability given by a strictly decreasing function \( \alpha_t \in [0, 1] \) in \( t \).
Applying masks to the tokens with probability \( 1 - \alpha_t\) results in a partially masked sequence \( x_t \). 
When \( t=1 \), all tokens are masked; when \( t=0 \), the sequence remains unchanged.

\subsection{Training}
\paragraph{Pre-training.}
The reverse process is parameterized by a mask predictor \( P_\theta(\cdot \mid x_t) \), which takes the partially masked sequence \( x_t \) as input and predicts all masked tokens in parallel. 
Let the indicator $\mathbf{1}_{[x_t^i = \textrm{M}]}$ be 1 if the $i$-th token in $x_t$, namely $x_t^i$, is masked (represented by $x_t^i=\textrm{M}$), and 0 otherwise.
Then, training proceeds by sampling $t$ from a uniform distribution $\mathcal{U}[0,1]$, i.e., \( t \sim \mathcal{U}[0,1] \), constructing \( x_t \) via the forward process, and minimizing a masked cross-entropy loss evaluated only at masked positions; thus, the pre-training objective is:
\begin{align}
\label{eq:mdlm-objective}
\mathcal{L}^{\text{\tiny CE}}_{\theta,t}
&\triangleq
- \mathbb{E}_{t, x_0, x_t}
\Biggr[
\gamma_t
\sum_{i=1}^{L}
\mathbf{1}_{[x_t^i = \textrm{M}]}
\log P_\theta(x_0^i \mid x_t)
\Biggr]
,
\end{align}
where $\gamma_t = -\frac{\alpha'_t}{1-\alpha_t}$, and $\alpha'_t$ represents the derivative of $\alpha_t$ with respect to $t$.\footnote{ See detailed explanation in~\cite{sahoo2024simple}.}

\paragraph{Supervised fine-tuning.}
SFT adapts a pre-trained MDLM to instruction-following using paired data \( (q_0, r_0) \), where \( q_0 \) denotes the prompt and \( r_0 \) the corresponding response. The objective is to model the conditional distribution \( P_\theta(r_0 \mid q_0) \).
Similar to pre-training, each token in \( r_0 \) is independently masked with probability \( \alpha_t \), as explained above, a monotonically decreasing function of \( t \), to obtain \( r_t \), while \( q_0 \) is kept intact. 
The mask predictor conditions on the concatenation of \( q_0 \) and \( r_t \) to predict masked response tokens. 
We then redefine the pre-training objective shown in Eq.~\eqref{eq:mdlm-objective} as the SFT objective as follows:
\begin{align}
\label{eq:sft-ce-objective}
\mathcal{L}^{\text{\tiny{CE}}}_{\theta,t}
&= 
- \mathbb{E}_{t, q_0, r_0, r_t}
\Biggr[
\gamma_t
\sum_{i=1}^{L_r}
\mathbf{1}_{[r_t^i = \textrm{M}]} %
\psi_{\theta}(r_0^i, q_0, r_t) 
\Biggr]
,
\end{align}
where \( L_r \) denotes the response length, and 
$\psi_{\theta}(r_0^i, q_0, r_t) = \log P_\theta(r_0^i \mid q_0, r_t)$

\paragraph{Relation between pre-training and SFT.}
Eq.~\eqref{eq:sft-ce-objective} can also cover the pre-training objective if we view pre-training input token sequences as never including a prompt part $q_0$ but only a response part; namely, $q_0$ is always an empty sequence and $x_t = r_t$.

\subsection{Inference}
\label{subsec:inference}

After training, text generation discretizes the reverse diffusion process to sample from the conditional distribution \( P_\theta(r_0 \mid q_0) \).
Following standard MDLM decoding (e.g., LLaDA), we fix the response length \( L_r \) as a hyperparameter and initialize generation of \( r_t \) at \( t = 1 \), yielding a fully masked response sequence, i.e.,
\begin{align}
r_{1} \!=\! (\texttt{<MASK>}, \ldots, \texttt{<MASK>}) \!\in\! (\mathcal{V} \!\cup\! \{\texttt{<MASK>}\})^{L_r}
,
\end{align}
where \( \mathcal{V} \) denotes the model's vocabulary.
At an intermediate timestep \( t \) (from $t=1$ to $t=0$), all masked tokens are predicted simultaneously and independently. After prediction, a subset of the newly revealed tokens is re-masked at each step to ensure that the overall masking ratio follows the prescribed noise schedule. Following LLaDA, the low-confidence remasking~\cite{chang2022maskgitmaskedgenerativeimage} is adopted; tokens with lower prediction confidence are selected for re-masking, and tokens that remain unmasked are carried forward and not re-masked in subsequent steps, making this process irreversible.

MDLMs' generation are known to be sensitive to inference setting such as generation length, generation steps, denoising schedule, and masking strategy~\cite{nie2025llada,zhu2025llada1p5}, the focus of this work is to demonstrate the influence of misplaced token could cause and the effectiveness of relaxing it, not to optimize for the best absolute score, therefore we adopt a fixed inference configuration across all experiments: response length $L=1024$, $N=512$ diffusion steps, and the standard low-confidence remasking strategy used in LLaDA. This setup ensures that any observed differences are attributable to the applied training changes rather than to decoding hyperparameters.

\section{Motivation: Sensitivity to Positional Misalignment}
\label{sec:motivation}

Standard MDLM training builds positional rigidity into the objective at two interfaces. First, each prediction is aligned to its target at a fixed output position. Second, the context that surrounds each masked token is assumed to be gold text at gold positions. In NAT, single-pass decoding means only the first interface is in play, and at that interface CE-trained decoders have been argued to be sensitive to small positional shifts~\cite{ghazvininejad2020axe}.

In contrast, MDLMs are exposed to both interfaces and decode iteratively. A small positional shift introduced at one step then enters the conditioning context for every later step, where it is treated as gold. We study two questions: whether such positional inconsistencies arise naturally during standard decoding, and whether introducing them causally degrades generation quality.

\subsection{Setup}

We run experiments on LLaDA-8B-Instruct~\cite{nie2025llada}, the most widely used open MDLM trained from scratch under the standard masked diffusion objective described in \cref{sec:background}, where supervision at position $i$ targets the token at position $i$ (cf. Eq.~\eqref{eq:sft-ce-objective}). This makes LLaDA a clean testbed for isolating the positional sensitivity that the masked diffusion objective itself induces\footnote{We leave out Dream~\cite{ye2025dream}, another commonly used open MDLM, because it is initialized from an autoregressive model and uses a shift-operation training scheme that supervises the hidden state at position $i$ to predict position $i{+}1$, carrying a left-to-right positional bias forward.}. We use the 750 prompts from Arena-Hard-v2.0~\cite{li2024crowdsourced,arenahard2024} and generate one standard, unintervened response trajectory for each prompt. The two studies below share only this setup: the first reads these trajectories directly, and the second intervenes on them and compares against them.

\subsection{Naturally Occurring Positional Regret}

\paragraph{Metric.}
We record the model's top-1 prediction at every response position and decoding step in the unintervened trajectories. Let token $a$ become permanently committed at position $i$ at step $c$, and let $j \in \{i-1,i+1\}$. We count an adjacent later-regret event when two conditions hold: (i) at step $c$, $a$ is the top-1 prediction at $i$ but not at $j$; and (ii) at the earliest qualifying later step $u>c$, $a$ is no longer top-1 at $i$ but is top-1 at $j$. To rule out ordinary adjacent repetitions, we require the final token at $j$ to differ from $a$, and we count each directed pair $(i,j)$ only once.

\paragraph{Results.}
Among the 750 unintervened trajectories, 342 responses (45.6\%) contain at least one adjacent later-regret event. On average, affected tokens are committed at 48.9\% of response progress, while the matching regret signal appears at 72.2\%, after much more surrounding context has been revealed. Thus, prediction patterns consistent with adjacent positional displacement arise naturally during standard MDLM decoding. This observational result, however, does not establish whether such displacement affects final generation quality.

\subsection{Effect of Small Positional Shifts}

\begin{wrapfigure}[17]{r}{0.5\columnwidth}
    \centering
    \includegraphics[width=0.48\columnwidth]{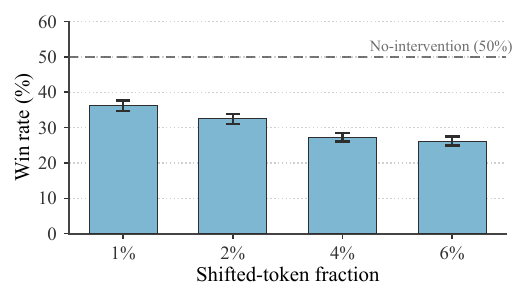}
    \caption{Win rate (\%) on Arena-Hard of intervened LLaDA-8B-Instruct generations \emph{against their non-intervened counterparts}. The dashed line at 50\% marks the no-intervention reference. Error bars indicate 95\% confidence intervals.}
    \label{fig:preliminary}
\end{wrapfigure}

\paragraph{Shift intervention.}
To test whether such positional displacements affect generation quality, we intervene at fixed time steps during decoding. At each such step, we uniformly sample $K$ locations where the current token has already been unmasked and a neighboring position is still \texttt{<MASK>}, and we swap each such pair. Because the swap only exchanges a revealed token with an adjacent masked slot, it keeps the revealed token itself; the only change is a one-position displacement. We apply this procedure every $0.05$ diffusion time units for $t \geq 0.5$, which yields 10 intervention events within the 512-step generation. Each event shifts exactly $K$ tokens, so in total $10 \times K$ out of 1024 tokens are intervened. We call this the \textit{shifted-token fraction}, and we sweep $K$ to vary the intervention strength.

\paragraph{Evaluation.}
For each prompt, we compare the intervened response against its unintervened counterpart, following the official Arena-Hard protocol and reporting the overall win rate. This paired comparison isolates the effect of the positional intervention rather than the absolute generation quality. We use GPT-5.4 as the judge model, and we decode deterministically to eliminate the judge variance that comes from the default configuration.

\paragraph{Results.}
Figure~\ref{fig:preliminary} reports win rates under the controlled shift intervention. Even when only a small fraction of generated tokens (1\%--6\%) is displaced by one position, the win rate falls well below the 50\% no-intervention reference. The effect appears at shifted-token fractions as low as 1\%, showing that MDLMs are sensitive to small positional shifts under iterative parallel decoding\footnote{Our trained model stays robust under the same intervention (\cref{sec:discussion}), which indicates that the drop is not caused simply by linguistic corruption from the token swap.}.

Together, these results show that positional inconsistencies arise during standard decoding, and that controlled cases of such displacement measurably harm generation quality. This motivates us to explore relaxing CE's positional rigidity on the training side, as a design lever for MDLM SFT. In the NAT literature, this sensitivity has been addressed by replacing position-wise CE with an alignment-flexible objective such as CTC~\cite{saharia2020ctcnat,libovicky2018ctcnat}. We adapt this approach to the MDLM SFT setting in \cref{sec:method}, with modifications that fit the MDLM training pipeline and preserve target surface forms.

\section{Method: Relaxing Positional Alignment}
\label{sec:method}
As motivated in \cref{sec:motivation}, we adapt a modified form of CTC~\cite{graves2006ctc} to the MDLM SFT setting as an auxiliary objective alongside the standard MDLM cross-entropy loss. 
Standard CTC allows the model to emit a longer latent alignment containing special blank symbols and repeated tokens, and recovers the target via a collapse map that removes blanks and merges adjacent duplicates. In our setting, however, merging adjacent duplicates can distort target surface forms, for example, collapsing ``**'' into ``*'' or ``100'' into ``10''. We therefore modify the collapse map to remove only blank symbols.

\subsection{Vanilla CTC objective}
Under the standard CTC conditional independence assumption, the alignment probability factorizes as:
\begin{align}
\label{eq:ctc-factorization}
P^{\text{\tiny CTC}}(a \mid q_0, \tilde r_t)
=
\prod_{i=1}^{\tilde L}
P_\theta(a^i \mid q_0, \tilde r_t).
\end{align}
The CTC likelihood of the clean target response $r_0$ is obtained by marginalizing over all alignments that collapse to $r_0$:
\begin{align}
\label{eq:ctc-marginal}
P^{\text{\tiny CTC}}_\beta(r_0 \mid q_0, \tilde r_t)
=
\sum_{a \in \beta^{-1}(r_0)}
P^{\text{\tiny CTC}}(a \mid q_0, \tilde r_t),
\end{align}
where $\beta$ is a collapse map and $\beta^{-1}(r_0)$ denotes the set of alignments that collapse to $r_0$ under $\beta$. Standard CTC uses $\beta_{\text{std}}$, which first replaces each maximal run of identical symbols in $a$ with a single symbol and then removes all \textsc{\textless slack\textgreater} symbols. 
The corresponding training objective is:
\begin{align}
\label{eq:sft-ctc-objective}
\mathcal{L}^{\text{\tiny CTC}}_{\theta,t}
=
- \mathbb{E}_{t,\, q_0,\, r_0,\, \tilde r_t}
\!\left[
\log P^{\text{\tiny CTC}}_{\beta_{\text{std}}}(r_0 \mid q_0, \tilde r_t)
\right].
\end{align}

\subsection{Modified CTC Objective}
\label{subsec:ctc-s}

Our method has two components: blank-augmented training data and a modified CTC objective.

\paragraph{Data augmentation.}
We construct the data using \textsc{\textless slack\textgreater} token as blank symbol needed in CTC object: $\tilde r_0 \in (\mathcal{V} \cup \{\textsc{\textless slack\textgreater}\})^{\tilde L}$ from the gold response $r_0 \in \mathcal{V}^{L}$ by inserting monotonic linear decreasing \textsc{\textless slack\textgreater} ratio with $t$, to match the monotonic increasing nature of \textsc{\textless slack\textgreater} tokens during the generation process. Specifically, we implement the \textsc{\textless slack\textgreater} ratio $s = s_{max}(1-t)$ with $s_{max} = 0.5$, so the slack ratio reaches at most $0.5$ at $t=0$, to prevent \textsc{\textless slack\textgreater} dominating the generated text.

\paragraph{CTC-S: \textsc{\textless slack\textgreater}-only collapse.}
As noted in Eq.~\eqref{eq:ctc-marginal}, the duplicate-merging step in $\beta_{\text{std}}$ can distort target surface forms (e.g., ``**''~$\to$~``*'', ``100''~$\to$~``10''), which is problematic for text generation. We therefore define a modified collapse map $\beta_{\text{slack}}$ that removes only \textsc{\textless slack\textgreater} symbols and preserves adjacent duplicates in $a$. Substituting $\beta_{\text{slack}}$ for $\beta_{\text{std}}$ in Eq.~\eqref{eq:ctc-marginal} yields our \textbf{CTC-S} objective:
\begin{align}
\label{eq:sft-ctcs-objective}
\mathcal{L}^{\text{\tiny CTC-S}}_{\theta,t}
=
- \mathbb{E}_{t,\, q_0,\, r_0,\, \tilde r_t}
\!\left[
\log P^{\text{\tiny CTC}}_{\beta_{\text{slack}}}(r_0 \mid q_0, \tilde r_t)
\right].
\end{align}
CTC-S retains the alignment flexibility of standard CTC while preserving target surface forms, and is the auxiliary objective we use throughout the rest of the paper\footnote{We provide algorithm on CTC-S objective in Appendix~\ref{appendix:ctcs_algo}}.

\subsection{Combined Objective}
\label{subsec:combined-objective}
We redefine the SFT objective to incorporate the CTC-S objective into SFT as follows:
\begin{align}
\label{eq:sft-slackdiffusion-objective}
\mathcal{L}^{\text{\tiny SFT}}_{\theta,t}
=
\mathcal{L}^{\text{\tiny CE}}_{\theta,t}
+
\lambda \mathcal{L}^{\text{\tiny CTC-S}}_{\theta,t},
\end{align}
where \( \lambda \) controls the contribution of the CTC-S term.
The CTC-S term is computed against the original slack-free target \(r_0\), as explained in Eq.~\eqref{eq:sft-ctc-objective}, whereas the CE term is computed over the slack-augmented target \(\tilde r_0\).
Moreover, both CTC-S and CE terms use the corrupted response sequence \(\tilde r_t\), which is derived from \(\tilde r_0\).
Therefore, when computing the CE term, we compute Eq.~\eqref{eq:sft-ce-objective} by substituting \(r_t\) and \(r_0\) with \(\tilde r_t\) and \(\tilde r_0\), respectively.

This encourages the model to emit \textsc{\textless slack\textgreater} tokens in \(\tilde L\) predictions when needed to accommodate positional uncertainty, while keeping the supervised output content equal to \(r_0\) after collapse.

\section{Experiments}
\label{sec:experiments}
To demonstrate the effectiveness of our method, we conduct experiments on four open-ended generation benchmarks.

\subsection{Experimental Setup}

\paragraph{Data preparation.}
\label{subsec:data_prepare}
We used the official filtered Magpie-Pro dataset~\cite{xu2025magpie}, which contains approximately 300k high-quality instruction-following examples synthesized from Llama-3-70B-Instruct~\cite{grattafiori2024llama3herdmodels}. To balance computational cost and the effect of the CTC-S objective, we further filter the dataset by response length, retaining examples with responses between 512 and 1024 tokens under the LLaDA tokenizer. This results in approximately 291k training samples.

\paragraph{Models.}
We train three models on top of LLaDA-8B-Instruct (LLaDA hereafter)~\cite{nie2025llada} to isolate the contributions of our method. 
The \emph{CE-Only} model is trained with the standard MDLM cross-entropy objective on our Magpie subset, serving as a baseline that controls for the effect of additional SFT data. 
The \emph{Jitter} model is trained with the same objective on the \textsc{\textless slack\textgreater}-augmented data, exposing the model to position-shifted contexts under hard position-wise supervision. 
The \emph{CE + CTC-S} model (CTC-S hereafter) is trained with the combined CE + CTC-S objective on the same slack-augmented data; this is the only model that trained with CTC-S. All other training settings are identical.\footnote{We describe the further implementation details in Appendix~\ref{appendix:imple_details}}.

\paragraph{Benchmarks.}
We evaluate open-ended generation quality on four benchmarks: Arena-Hard-v2.0~\cite{li2024crowdsourced,arenahard2024}, MT-Bench~\cite{zheng2023judging}, WildBench~\cite{lin2025wildbench}, and Creative Writing Bench v3~\cite{creative-writing-bench-v3}.
Note that for Arena-Hard-v2.0, it uses relative win rates against a predefined baseline. We replace the standard o3 baseline with CE-Only model, as LLaDA achieve near-zero win rates against o3, leaving no signal to discriminate among variants. 
Results are intended for controlled comparison in this work and are not leaderboard-comparable.

\paragraph{Metrics.}
For Arena-Hard, we include the bootstrapped 95\% confidence interval following the official setting.
Additionally for each benchmark, we performed paired-bootstrap comparisons between every pair of models on the evaluation instances scored for both models. We resampled the aligned instance-level score differences with replacement for 10,000 bootstrap iterations and computed the mean score difference and its 95\% confidence interval. 
We regard a comparison as statistically significant when the confidence interval excludes zero\footnote{Full pairwise confidence intervals are reported in Appendix~\ref{appendix:significance}}.

\begin{table}[t]
\centering
\setlength{\tabcolsep}{4.2pt}
\caption{Performance comparison on open-ended text generation benchmarks. For Arena-Hard, we report bootstrapped 95\% confidence intervals following the official setting. $\dagger$ and $\ddagger$ indicate that CE + CTC-S is significantly better than the CE-Only model and the Jitter model, respectively (paired bootstrap, $p<0.05$). Bold indicates the best score in each column.}

\newcommand{\scoreci}[3]{%
$#1$\ {\scriptsize($#2/#3$)}%
}
\begin{tabular}{lcccc}
\toprule
& \textbf{Arena-Hard\(^*\)}
& \textbf{Creative-Writing-Bench v3}
& \textbf{MT-Bench}
& \textbf{Wild-Bench} \\
\cmidrule(lr){1-1} \cmidrule(lr){2-2} \cmidrule(lr){3-3} \cmidrule(lr){4-4} \cmidrule(lr){5-5}
\textbf{LLaDA}
& \scoreci{43.7}{-1.1}{+1.4}
& $23.2$
& $2.84$
& $-5.48$ \\
\midrule
\textbf{CE-Only}
& \scoreci{50.0}{-0.0}{+0.0}
& $24.6$
& $3.44$
& $-4.53$ \\
\textbf{Jitter}
& \scoreci{48.4}{-1.4}{+1.5}
& $25.1$
& $3.61$
& $-4.46$ \\
\textbf{CE + CTC-S}
& \scoreci{\mathbf{55.1}^{\dagger\ddagger}}{-1.7}{+1.3}
& $\mathbf{27.7}^{\dagger\ddagger}$
& $\mathbf{4.35}^{\dagger\ddagger}$
& $\mathbf{-4.23}^{\dagger\ddagger}$ \\
\bottomrule
\end{tabular}

\label{tab:main}
\end{table}

\subsection{Results}
\paragraph{Overall.}
Table~\ref{tab:main} reports results on all tested benchmarks. We read the table as a three-step ladder, where each step isolates one factor:
(i) additional SFT data (CE-Only vs.\ LLaDA),
(ii) data augmentation under hard alignment (Jitter vs.\ CE-Only), and
(iii) flexible positional alignment (CTC-S vs.\ Jitter).
At a high level, CE-Only consistently outperforms the original LLaDA model, confirming that the training pipeline and data are healthy; and CTC-S achieves the best score on every benchmark, with significant gains over both Jitter ($\ddagger$) and CE-Only ($\dagger$) on all four benchmarks.

\begin{wrapfigure}[21]{r}{0.5\columnwidth}
    \centering
    \vspace{-0.5em}
    \includegraphics[width=0.48\columnwidth]{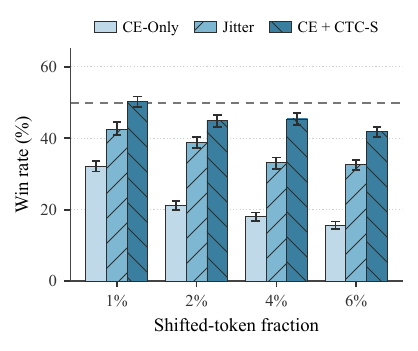}
    \caption{Robustness check under the controlled shift intervention. Win rate (\%) on Arena-Hard of intervened model generations against their non-intervened counterparts. The dashed line at 50\% marks the no-intervention reference. Error bars indicate 95\% confidence intervals calculated following the standard Arena-Hard setting.}
    \label{fig:ctc-robust}
\end{wrapfigure}

\paragraph{Contribution of \textsc{\textless slack\textgreater} augmentation.}
Comparing the Jitter model and the CE-Only model isolates the effect of \textsc{\textless slack\textgreater} augmentation in our method. The Jitter model's results are mixed: on Arena-Hard, its confidence interval almost overlaps with that of the CE-Only model, while it shows moderate gains on the remaining benchmarks. 
One possible reading is that random \textsc{\textless slack\textgreater} tokens act primarily as a source of variation on the CE objective: the \textsc{\textless slack\textgreater} augmentation exposes the model to position-shifted contexts, but the objective itself still imposes hard position-wise alignment. We do not test this directly.

\paragraph{Contribution of CTC-S objective.}
Further comparing the CTC-S model and the Jitter model isolates the effect of alignment-flexible objective. The CTC-S model is significantly better than the Jitter model ($\ddagger$) on all four tested benchmarks. This suggests that the alignment-flexible objective itself contributes to the improvement on open-ended generation, beyond what the slack-augmented data alone provides.

\paragraph{Takeaway.}

Across the three runs, the gain is concentrated at the CTC-S step, the only step that introduces alignment flexibility into the objective. This identifies training-side alignment flexibility as a useful design dimension for MDLM SFT\footnote{Appendix~\ref{appendix:llada-moe} reports the same experiment on LLaDA-MoE.}, complementary to the inference-time interventions explored in prior work.

\section{Discussion}
\label{sec:discussion}
Beyond the main results in \cref{sec:experiments}, several natural follow-up questions remain. (i) Since CTC-S targets the positional shift sensitivity, does it deliver the robustness property it was designed for? (ii) How does the CTC-S model use the <SLACK> token during generation? (iii) Does relaxing positional alignment come at the cost of general capabilities? (iv) How sensitive is the method to the maximum slack ratio $s_{\max}$? We address these in turn\footnote{We also provide a transferability check on LLaDA-MoE and a comparison with inference-time block diffusion in Appendix~\ref{appendix:llada-moe} and Appendix~\ref{appendix:block-diffusion}.}.

\subsection{Robustness check under the controlled shift intervention}
CTC-S targets robustness to the controlled shift intervention from §\ref{sec:motivation}, which we test here.
We repeat the intervention for the CE-Only, Jitter, and CTC-S models; Figure~\ref{fig:ctc-robust} summarizes the results.

The CE-Only model shows nearly the same sensitivity as the original LLaDA (Figure~\ref{fig:preliminary}).
Jitter sits below the 50\% no-intervention reference at all tested fractions; this indicates that slack augmentation closes the gap by a large margin, though not entirely.
CTC-S substantially more robust than CE-Only and Jitter,
therefore exhibits the robustness property it was designed for.

\subsection{\textsc{\textless slack\textgreater} Token Behavior}
We examine \textsc{\textless slack\textgreater} usage in the CTC-S model's MT-Bench generations along two axes: overall emission rate and positional distribution.

\paragraph{\textsc{\textless slack\textgreater} usage is selective across prompts.} Although every training response contains \textsc{\textless slack\textgreater} tokens, only 83 of 160 MT-Bench samples (51.9\%) emit any \textsc{\textless slack\textgreater} at inference. This suggests the model learned to treat \textsc{\textless slack\textgreater} as an optional resource rather than a required output feature.

\paragraph{\textsc{\textless slack\textgreater} emission rate is close to the training augmentation rate.} Among samples that emit \textsc{\textless slack\textgreater}, it accounts for 27.5\% of generated tokens on average, comparable to the 25\% per-sample augmentation rate used during training. This suggests the model did not drift toward over- or under-emission relative to its training signal.

\paragraph{\textsc{\textless slack\textgreater} is mainly allocated to inter-word positions.}
\textsc{\textless slack\textgreater} is injected at random positions during training, including inside subword sequences. At inference, however, only 4.9\% of emitted \textsc{\textless slack\textgreater} tokens fall at subword-internal positions\footnote{A \textsc{\textless slack\textgreater} position is classified as subword-internal when the following token does not begin with a leading space and is not punctuation; sequence-initial positions are excluded.}. Without explicit supervision, the model concentrates \textsc{\textless slack\textgreater} at inter-word boundaries rather than within subwords. 
This is consistent with the alignment-flexibility view: linguistically, positional uncertainty is most meaningful between words, where multiple lexicalizations and orderings are possible, but not within a word, since subword pieces are an artifact of tokenization rather than linguistic units and the boundaries between them carry no independent meaning.

\subsection{General-Capability Benchmarks}
\label{subsec:general-capability}
\begin{wraptable}[15]{r}{0.5\columnwidth}
\centering
\caption{Performance on general capability benchmarks. Bold indicates the best score among models trained on our dataset (CE-Only, Jitter, CTC-S); the LLaDA column is included as a reference. $^{\ast}$ indicates scores taken from the official repository.}
\small
\setlength{\tabcolsep}{3pt}

\begin{tabular}{lcccc}
\toprule
& \multicolumn{3}{c}{\textbf{Trained}}
& \textbf{Reference} \\
\cmidrule(lr){2-4} \cmidrule(lr){5-5}
& \textbf{CE-Only}
& \textbf{Jitter}
& \textbf{CTC-S}
& \textbf{LLaDA} \\
\cmidrule(lr){1-1} \cmidrule(lr){2-2} \cmidrule(lr){3-3} \cmidrule(lr){4-4} \cmidrule(lr){5-5}
\textbf{GPQA}
& $29.8$ & $29.8$ & $\mathbf{31.3}$ & $32.3^{\ast}$ \\
\textbf{MMLU}
& $62.1$ & $\mathbf{63.8}$ & $63.4$ & $65.4^{\ast}$ \\
\textbf{MBPP}
& $36.2$ & $\mathbf{37.2}$ & $36.8$ & $39.6^{\ast}$ \\
\textbf{IFEval}
& $63.2$ & $65.0$ & $\mathbf{66.9}$ & $65.2^{\ast}$ \\
\textbf{GSM8K}
& $66.1$ & $\mathbf{69.9}$ & $68.9$ & $68.8^{\ast}$ \\
\textbf{MATH500}
& $26.6$ & $\mathbf{27.2}$ & $26.0$ & $31.4$ \\
\bottomrule
\end{tabular}

\label{tab:general-capability}
\end{wraptable}

To confirm that CTC-S does not harm capabilities beyond open-ended generation, we report results on GPQA \cite{rein2024gpqa}, MMLU \cite{hendrycks2021mmlu}, MBPP \cite{austin2021mbpp}, IFEval \cite{zhou2023ifeval}, GSM8K \cite{cobbe2021gsm8k}, and MATH500 \cite{hendrycks2021meath,lightman2024letsverify}. We use LLaDA's official OpenCompass evaluation and match its generation and evaluation settings.\footnote{See \url{https://github.com/ML-GSAI/LLaDA/blob/main/EVAL.md} for details.} 
The CTC-S vs.\ CE-Only comparison isolates the effect of our objective from the effect of additional SFT data.

\paragraph{Capability regression is data-driven, not objective-driven.}
All three trained models regress modestly relative to LLaDA on GPQA, MMLU, MBPP, and MATH500 (\Cref{tab:general-capability}); the regression is shared across variants, indicating that it stems primarily from the SFT data rather than the CTC-S objective, consistent with the Magpie paper's observation~\cite{xu2025magpie} that Magpie-Pro underperforms on reasoning- and coding-heavy benchmarks.

\paragraph{CTC-S preserves base capabilities where the data permits.}
CTC-S matches or slightly exceeds the LLaDA reference on both (66.9 vs.\ 65.2 on IFEval; 68.9 vs.\ 68.8 on GSM8K). Relative to CE-Only, CTC-S matches or improves on five of the six benchmarks, with a marginal gap only on MATH500 (26.0 vs.\ 26.6). Relaxing positional alignment therefore does not come at the cost of broader capabilities: on the two benchmarks where the SFT data does not impose a ceiling, CTC-S preserves the base model's performance.

\subsection{Slack-Ratio Ablation}
\label{subsec:slack-ratio}
To test how sensitive our method is to the slack ratio $s_{\max}$, we train two additional CE~+~CTC-S variants with $s_{\max}=0.2$ and $s_{\max}=0.8$ under the same training and evaluation setup as \S\ref{sec:experiments}.

\begin{table}[t]
\small
\centering
\setlength{\tabcolsep}{4.5pt}
\caption{Slack-ratio ablation results. For Arena-Hard, we report bootstrapped 95\% confidence intervals following the official setting. $\dagger$/\S\ indicate significantly better/worse than the CE-Only reference (paired bootstrap, $p<0.05$). Bold indicates the best score in each column.}
\newcommand{\scoreci}[3]{%
$#1$\ {\scriptsize($#2/#3$)}%
}
\begin{tabular}{lcccc}
\toprule
& \textbf{Arena-Hard\(^*\)}
& \textbf{Creative-Writing-Bench v3}
& \textbf{MT-Bench}
& \textbf{Wild-Bench} \\
\cmidrule(lr){1-1} \cmidrule(lr){2-2} \cmidrule(lr){3-3} \cmidrule(lr){4-4} \cmidrule(lr){5-5}
\textbf{CE-Only (reference)}
& \scoreci{50.0}{-0.0}{+0.0}
& $24.6$
& $3.44$
& $-4.53$ \\
\midrule
\textbf{CE + CTC-S}, $s_{\max}=0.2$
& \scoreci{45.1^{\text{\S}}}{-1.3}{+1.5}
& $27.31^{\dagger}$
& $3.96^{\dagger}$
& $-4.50$ \\
\textbf{CE + CTC-S}, $s_{\max}=0.5$
& \scoreci{\mathbf{55.1}^{\dagger}}{-1.7}{+1.3}
& $\mathbf{27.70}^{\dagger}$
& $4.35^{\dagger}$
& $-4.23^{\dagger}$ \\
\textbf{CE + CTC-S}, $s_{\max}=0.8$
& \scoreci{49.9}{-1.5}{+1.7}
& $27.68^{\dagger}$
& $\mathbf{4.41}^{\dagger}$
& $\mathbf{-4.15}^{\dagger}$ \\
\bottomrule
\end{tabular}

\label{tab:slack-ratio}
\end{table}

Table~\ref{tab:slack-ratio} shows that the method remains effective across this range: every setting improves significantly over CE-Only on Creative-Writing-Bench and MT-Bench. $s_{\max}=0.5$ gives the best overall balance, while $s_{\max}=0.8$ scores slightly higher on MT-Bench and WildBench. The only significant regression in the table is at the smallest setting, where $s_{\max}=0.2$ falls below CE-Only on Arena-Hard; neither of the larger settings is significantly worse than CE-Only on any benchmark.

\section{Related Work}
\label{sec:related_work}
\subsection{Prior Approaches to MDLM Decoding Failures}
Prior work has identified several failure modes in the fixed-length, position-wise decoding setup of MDLMs and proposed a range of remedies. One line enables variable-length generation through token-position adjustment, end-of-sequence prediction, or insertion/deletion operations, addressing failures arising from fixed canvases and uncertain span boundaries~\cite{zhang2025ddot, wu2025dreamon, yang2025diffusionllmnativevariable, kim2025any, havasi2025editflows, li2025beyond}. A second line improves which tokens to commit at each step, through position-aware or guidance-based samplers and learned unmasking-order strategies~\cite{patel2025improved, huang2025pc, lee2025lookaheadunmaskingelicitsaccurate, jazbec2025learningunmaskingpoliciesdiffusion}. A third line targets early commitment by allowing already-revealed tokens to be revised, either at inference through re-masking and refinement~\cite{wang2025remasking, peng2025pathplanningmaskeddiffusion} or through training-side objectives that teach self-correction~\cite{huang2025dontsettleearlyselfreflective, kim2025finetuningmaskeddiffusionprovable, zhang2025correctivediffusionlanguagemodels}.

Our work differs in what it changes: rather than adjusting the canvas, the unmasking order, or the model's ability to revise committed tokens, we relax position-wise supervision during training, keeping the standard MDLM decoding setup unchanged.

\subsection{Non-autoregressive Translation}
\label{sec:related-nat}
Non-autoregressive translation \citep{gu2018nat} replaces left-to-right decoding with parallel prediction under a conditional independence assumption. Because a single source has many valid translations, a parallel decoder trained with position-wise cross-entropy receives gradient signal that mixes incompatible modes at the same position. Position-wise CE penalizes a correct token at the wrong position as harshly as a wrong token, a sensitivity to positional shifts that the NAT literature has repeatedly identified \citep{ghazvininejad2020axe,du2021order,du2022ngram}. MDLMs inherit this setup, which motivates studying the same sensitivity.

NAT has developed a family of training-side responses to relax this constraint. CTC is one established line: first applied to NAT \citep{libovicky2018ctcnat}, paired with an iterative variant~\cite{chan2020imputer} to close the gap to AR baselines on WMT \citep{saharia2020ctcnat}, and identified as an essential ingredient for a competitive fully-NAT system \citep{gu2021fully}. Other approaches reach similar gains through best-alignment objectives, marginalization over alignment paths, or curricula that progressively reveal target tokens \citep{du2021order,du2022ngram,huang2022directed,qian2021glancing}.

Our work transfers this training-side ingredient to MDLM SFT through CTC, whose blank-augmented alignment space and forward--backward marginalization let it absorb positional uncertainty within MDLMs' fixed decoding canvas.

\subsection{Alignment-Flexible Objectives and Structured Supervision}
\label{sec:related-ctc}
CTC \citep{graves2006ctc} originated as a training criterion for sequence labeling with unobserved input--output alignment, most prominently in speech recognition \citep{graves2012sequence,graves2013speech}. 
More broadly, recent work has highlighted the importance of task-specific supervision and representation design \citep{shen2025fine,yu2026elsa3d,yu2025core3d,
liu2026palm,li2026toward}.
Other alignment-flexible objectives also includes stochastic edit-distance models \citep{ristad1998learning,oncina2006stochastic}, semi-Markov conditional random fields \citep{sarawagi2004semimarkov}, and differentiable dynamic time warping \citep{cuturi2017softdtm}. Within this class, CTC is distinctive in marginalizing over alignments via a blank-augmented label space and a deterministic collapse map, admitting an efficient forward--backward dynamic program. Later work combines the marginalized objective with iterative refinement \citep{higuchi2020maskctc} and extends CTC to non-speech sequence transduction, including non-autoregressive speech translation \citep{xu2023ctcbased} and latent-variable extensions \citep{fujita2024lvctc}.

Our method adapts the canonical CTC objective to text generation.

\section{Conclusion}
\label{sec:conclusion}
This work studies MDLMs through the lens of non-autoregressive translation (NAT): both train parallel decoders with position-wise cross-entropy, and CE-trained parallel decoders have been argued to be sensitive to small positional shifts in the NAT setting. Through a controlled shift intervention on LLaDA-8B-Instruct, we find that displacing as little as 1\% of generated tokens by a single position substantially reduces win rates on Arena-Hard, indicating that MDLMs share this sensitivity under iterative parallel decoding. Motivated by this, we adapt connectionist temporal classification (CTC), an alignment-flexible objective known to mitigate this sensitivity in NAT, to MDLM supervised fine-tuning. We introduce CTC-S, a modified form of CTC that uses a special \textsc{\textless slack\textgreater} token to absorb positional uncertainty between target tokens and output positions and an adjusted collapse map that preserves target surface forms. The resulting model consistently improves over both the original model and a matched cross-entropy baseline across four open-ended generation benchmarks, while preserving performance on standard capability benchmarks.

These results identify training-side alignment flexibility as a useful design dimension for MDLM supervised fine-tuning, complementary to the inference-time interventions explored in prior work. We hope this NAT--MDLM bridge encourages further work on flexible alignment objectives in MDLM training, including extension to pre-training and to other instantiations of alignment flexibility.

\paragraph{Limitations.}
The \textsc{\textless slack\textgreater} mechanism is one instantiation of alignment flexibility; other formulations, including non-CTC objectives and learned slack budgets, may offer further gains. Our experiments are limited to SFT stage, leaving the effect of relaxing positional alignment during pre-training unexplored. 
Our method operates at the training-objective level and is model-agnostic, verifying it on MDLMs trained under different schemes is left to future work.
We also fix the CTC-S loss weight $\lambda$ without tuning; sweeping it may yield further gains. 

\paragraph{Impact statement.} 
This work is methodological: we modify the SFT objective for masked diffusion language models and evaluate on standard open-ended generation benchmarks. To the extent that our approach narrows the open-ended generation gap between MDLMs and autoregressive models, it may contribute to MDLMs becoming a more practical alternative for general-purpose text generation, and inherits the broader societal considerations that apply to instruction-tuned LLMs. Our method does not introduce new capabilities beyond those of the base LLaDA-8B-Instruct model.

\paragraph{Ethics statement.} 
Our experiments use only publicly released artifacts: the LLaDA-8B-Instruct model and the Magpie-Pro dataset, the latter synthesized from Llama-3-70B-Instruct. We did not collect data from human subjects, and our training data and base model inherit whatever biases are present in these upstream resources. To promote transparency and reproducibility, we will release all code used in our experiments. Comprehensive details of our experimental setup are provided in each section and the appendix to ensure reproducibility.

\clearpage
\section*{Acknowledgments.}
We want to express our gratitude to the members of the Tohoku NLP Group for their insightful comments.
We used ABCI 3.0 provided by AIST and AIST Solutions with support from ``ABCI 3.0 Development Acceleration Use'' for our main experiments.
A part of the computation was carried out using the computer resource offered under the category of General Projects by Research Institute for Information Technology, Kyushu University.
This work was supported by the ``R\&D Hub Aimed at Ensuring Transparency and Reliability of Generative AI Models'' project of the Ministry of Education, Culture, Sports, Science and Technology; JST Moonshot R\&D Grant Number JPMJMS2011-35 (fundamental research); JSPS KAKENHI Grant Numbers JP25KJ0615 and JP26H02511; JST BOOST, Japan Grant Number JPMJBS2421.

\paragraph{Competing interests.} We declare no competing interests.

\bibliographystyle{abbrvnat}
\bibliography{references}

\newpage
\appendix
\onecolumn

\section{CTC-S Forward Pass}
\label{appendix:ctcs_algo}

This section describes how we compute the CTC-S loss $\mathcal{L}^{\text{\tiny CTC-S}}_{\theta,t}$ defined in Eq.~\eqref{eq:sft-ctcs-objective} via a modified forward recurrence. The procedure parallels the standard CTC forward--backward of \cite{graves2006ctc}; we describe the forward pass in Algorithm~\ref{alg:ctcs}, and the backward pass is symmetric.

Given a clean target $r_0 \in \mathcal{V}^{L}$, we construct an extended label sequence of length $2L+1$ by inserting a \textsc{\textless slack\textgreater} symbol before, between, and after every target token. States at even indices $s \in \{0, 2, \ldots, 2L\}$ correspond to \textsc{\textless slack\textgreater} positions; states at odd indices $s \in \{1, 3, \ldots, 2L{-}1\}$ correspond to the $\lceil s/2 \rceil$-th target token. The model emits log-probabilities $\log P[t, \cdot]$ over the vocabulary at each response position $t \in \{1, \ldots, \tilde L\}$, and the forward variable $\alpha[t, s]$ accumulates the log-likelihood of all prefixes of length $t$ that end in state $s$.

The algorithm differs from the standard CTC forward pass in exactly two transition rules, marked \textbf{M1} and \textbf{M2}:

\paragraph{M1 (self-loop restricted to \textsc{\textless slack\textgreater} states).}
Standard CTC permits a self-loop $\alpha[t{-}1, s] \!\to\! \alpha[t, s]$ at every state, which combined with the duplicate-merging collapse lets a single target token be emitted over multiple consecutive positions. Under $\beta_{\text{slack}}$, duplicates are not merged, so a self-loop at a label state would let the alignment emit the same target token at multiple adjacent positions. We therefore restrict self-loops to even (\textsc{\textless slack\textgreater}) states; label states must advance at every step.

\paragraph{M2 (unconditional skip at label states).}
Standard CTC permits the skip transition $\alpha[t{-}1, s{-}2] \!\to\! \alpha[t, s]$ only when the label at state $s$ differs from the label at state $s{-}2$, precisely to prevent two identical target tokens from being merged into one via the duplicate-collapse rule. Since $\beta_{\text{slack}}$ preserves duplicates, this guard is unnecessary: skipping the intervening \textsc{\textless slack\textgreater} state is always safe. We therefore allow the skip transition unconditionally at odd (label) states with $s \geq 2$.

Together, M1 and M2 ensure that the alignments reachable by the recurrence are exactly those in $\beta_{\text{slack}}^{-1}(r_0)$. The forward direction is immediate from the transition rules. For the reverse direction, any alignment $a$ with $\beta_{\text{slack}}(a) = r_0$ admits a unique decomposition into the three transition types (self-loop at \textsc{\textless slack\textgreater} states, advance, skip-at-label), since at each position the token $a_t$ and the current label index together determine the transition. The termination step marginalizes over the two valid final states ($s = 2L$, ending in \textsc{\textless slack\textgreater}, and $s = 2L-1$, ending in the final target token), yielding $\log P^{\text{CTC}}_{\beta_{\text{slack}}}(r_0 \mid q_0, \tilde{r}_t)$ from Eq.~~\eqref{eq:ctc-marginal}.

\begin{algorithm}[t]
\caption{CTC-S forward pass per example.}
\label{alg:ctcs}
\begin{algorithmic}[1]
\Require Log-probabilities $\log P \in \mathbb{R}^{\tilde L \times |\mathcal{V}|}$ at response positions, where $\log P[t, v] = \log P_\theta(v \mid q_0, \tilde r_t)$; clean target $r_0 \in \mathcal{V}^{L}$.
\Ensure CTC-S loss $\mathcal{L}^{\text{\tiny CTC-S}}$.
\State $\ell \gets (\textsc{\textless slack\textgreater},\, r_0^1,\, \textsc{\textless slack\textgreater},\, r_0^2,\, \ldots,\, r_0^{L},\, \textsc{\textless slack\textgreater})$ \Comment{Extended label sequence of length $2L{+}1$}
\State $\alpha[1, 0] \gets \log P[1, \textsc{\textless slack\textgreater}]$ \Comment{Initialization at $t=1$}
\State $\alpha[1, 1] \gets \log P[1, r_0^1]$ if $L > 0$, else $-\infty$
\State $\alpha[1, s] \gets -\infty$ for $s \in \{2, \ldots, 2L\}$
\For{$t = 2$ to $\tilde L$} \Comment{Forward recurrence}
  \For{$s = 0$ to $2L$}
    \State $a_{\text{self}} \gets \alpha[t{-}1, s]$ if $s$ even, else $-\infty$ \Comment{M1}
    \State $a_{\text{adv}} \gets \alpha[t{-}1, s{-}1]$ if $s \geq 1$, else $-\infty$
    \State $a_{\text{skip}} \gets \alpha[t{-}1, s{-}2]$ if $s$ odd and $s \geq 2$, else $-\infty$ \Comment{M2}
    \State $\alpha[t, s] \gets \textsc{logsumexp}(a_{\text{self}},\, a_{\text{adv}},\, a_{\text{skip}}) + \log P[t, \ell_s]$
  \EndFor
\EndFor
\State $\mathcal{L}^{\text{\tiny CTC-S}} \gets -\textsc{logsumexp}\bigl(\alpha[\tilde L, 2L],\, \alpha[\tilde L, 2L{-}1]\bigr)$ \Comment{Termination}
\State \Return $\mathcal{L}^{\text{\tiny CTC-S}}$
\end{algorithmic}
\end{algorithm}

\section{Implementation Details}
\label{appendix:imple_details}

\subsection{Overall Framework}
\label{subsec:overall_framework}

We perform SFT using a modified version of the training code from \cite{zhao2025d1}. The CE-Only baseline is trained with the standard diffusion cross-entropy objective on prompt/response pairs from our Magpie-Pro subset. The Jitter and CE + CTC-S models share an identical data pipeline that augments responses with \textsc{\textless slack\textgreater} tokens at training time and differ only in the loss function: Jitter applies the standard cross-entropy objective to the slack-augmented inputs, while CE + CTC-S adds the auxiliary CTC-S term defined in Eq.~\eqref{eq:sft-ctcs-objective}.

\paragraph{Data preparation.}
We use Magpie-Pro-300K-Filtered tokenized with the LLaDA tokenizer, discard examples whose response length falls outside $[512, 1024]$, and pad each example to a fixed length of 2048 tokens. For each example we retain the original response as the CTC-S target, before any \textsc{\textless slack\textgreater} insertion.

\paragraph{\textsc{\textless slack\textgreater} insertion.}
At training time, given a per-example timestep $t \sim \mathcal{U}[0, 1]$, we insert $n = \min\bigl(\lfloor s_{\max}(1-t)\,R \rfloor,\, A\bigr)$ \textsc{\textless slack\textgreater} tokens at uniformly sampled positions in the response, where $R$ is the response length, $s_{\max} = 0.5$, and $A$ is the available capacity within the 2048-token budget. The trailing \texttt{<EOT>} and \texttt{<EOS>} are excluded from \textsc{\textless slack\textgreater} insertion.

\paragraph{Loss computation.}
The cross-entropy term follows the standard MDLM formulation. 
The CTC-S term is computed on the response slice between the response start index and the end of the augmented response, against the slack-free clean response retained before augmentation, via the recurrence in Algorithm~\ref{alg:ctcs}.
The final training objective is the sum of the two terms with the CTC-S weight set to $\lambda = 0.1$, as in Eq.~\eqref{eq:sft-slackdiffusion-objective}.

\subsection{Training Hyperparameters}
\label{subsec:training_hyperparameters}

All three models are trained under the same optimization settings. We use LoRA adapters with rank $r=128$, scaling factor $\alpha=32$, and dropout $0.05$, without bias adaptation. For LLaDA-8B-Instruct, adapters are applied to the transformer projection layers $\{\texttt{q\_proj}, \texttt{k\_proj}, \texttt{v\_proj}\}$, the token embedding layer (\texttt{wte}) and the output projection of the feed-forward block (\texttt{ff\_out}).

Optimization uses AdamW~\cite{loshchilov2018decoupled} with a learning rate of $2\times10^{-4}$, weight decay $0.1$, and gradient clipping at $1.0$, under bfloat16 precision. Training is conducted for three epochs with a per-device batch size of 3, no gradient accumulation, and a maximum sequence length of 2048.
After length filtering, the dataset contains approximately 291k training examples. With distributed batching and batch-processing details, three epochs correspond to approximately 36k optimizer update steps.
We use a linear learning rate schedule without warmup and fix the random seed to 42. Checkpoints are evaluated and saved every 10\% of the training steps, and the checkpoint with the lowest evaluation loss is selected. We do not perform any hyperparameter search in this work.

\paragraph{LoRA targets for LLaDA-MoE.}
The transferability experiment in Appendix~\ref{appendix:llada-moe} repeats the same three-model comparison on LLaDA-MoE. The LoRA configuration mirrors the LLaDA setting (rank, scaling factor, dropout, optimizer, schedule, and seed are all unchanged), but the target-module names are remapped to the LLaDA-MoE module naming: $\{\texttt{q\_proj}, \texttt{k\_proj}, \texttt{v\_proj}\}$, the token embedding (\texttt{embed\_tokens}), and the language-model head (\texttt{lm\_head}). The router and expert MLPs are intentionally excluded so that the comparison isolates positional supervision rather than the MoE routing dynamics.

\subsection{Assets Used}
\label{subsec:other_details}
Table~\ref{tab:assets} lists all existing assets used in this work, together with their sources, licenses, and citations. We use the LLaDA-8B-Instruct and LLaDA-MoE-7B-A1B-Instruct checkpoints as base models, Magpie-Pro-300K-Filtered~\cite{xu2025magpie} as the SFT dataset, and a modified version of the d1 codebase as our training framework. For evaluation, we use the four open-ended generation benchmarks reported in \cref{sec:experiments} (Arena-Hard-Auto v2.0, Creative-Writing-Bench v3, MT-Bench, and WildBench), and the six capability benchmarks reported in \S\ref{subsec:general-capability} (GPQA, MMLU, MBPP, IFEval, GSM8K, and MATH500), the latter run through the LLaDA-forked OpenCompass evaluation pipeline. All assets are used in accordance with their published licenses.

\section{Pairwise Significance Test Results}
\label{appendix:significance}

To support the significance markers in Tables~\ref{tab:main}, \ref{tab:main_moe}, and \ref{tab:block_sweep_ablation}, we report the full pairwise 95\% paired-bootstrap confidence intervals for all relevant model pairs on aligned evaluation instances. For each benchmark and model pair, we compute instance-level score differences on the subset of examples scored for both models, resample those aligned differences with replacement for 10{{,}000 bootstrap iterations, and report the empirical 2.5th and 97.5th percentiles of the bootstrap mean difference (row minus column for the dense and MoE matrices; CE+CTC-S minus the column baseline at its best block size for the block-diffusion table). We regard a comparison as statistically significant when the confidence interval excludes zero.

Table~\ref{tab:table1_significance} corresponds to Table~\ref{tab:main} (LLaDA-8B-Instruct), Table~\ref{tab:main_moe_significance} to Table~\ref{tab:main_moe} (LLaDA-MoE), and Table~\ref{tab:block_sweep_significance} to Table~\ref{tab:block_sweep_ablation} (block-diffusion sweep ablation).

\section{Transferability to LLaDA-MoE}
\label{appendix:llada-moe}
We report the open-ended generation comparison from \S\ref{sec:experiments} on a second model, LLaDA-MoE, as a transferability check\footnote{The training setup matches \S\ref{sec:experiments}; the LoRA target-module remapping for LLaDA-MoE is described in Appendix~\ref{appendix:imple_details}.}.
Table~\ref{tab:main_moe} reports the results.
CE~+~CTC-S achieves the best score on three of four benchmarks and significantly improves over CE-Only ($\dagger$) on all four. The improvement over Jitter ($\ddagger$) is significant on Arena-Hard, Creative-Writing-Bench~v3, and WildBench.
On MT-Bench, CE~+~CTC-S is significantly better than CE-Only; the difference from Jitter is not significant, and Jitter has the highest score on this benchmark.

The overall pattern observed on LLaDA-8B-Instruct carries over: relaxing positional alignment via CTC-S consistently improves open-ended generation on a second MDLM trained under the same standard objective. Full pairwise confidence intervals are reported in Table~\ref{tab:main_moe_significance}.

\begin{table}[t]
\small
\centering
\setlength{\tabcolsep}{4.5pt}
\caption{Performance comparison on open-ended text generation benchmarks on LLaDA-MoE. For Arena-Hard, we report bootstrapped 95\% confidence intervals following the official setting. $\dagger$ and $\ddagger$ indicate that CE + CTC-S is significantly better than the CE-Only model and the Jitter model, respectively (paired bootstrap, $p<0.05$). Bold indicates the best score in each column.}
\newcommand{\scoreci}[3]{%
$#1$\ {\scriptsize($#2/#3$)}%
}
\begin{tabular}{lcccc}
\toprule
& \textbf{Arena-Hard\(^*\)}
& \textbf{Creative-Writing-Bench v3}
& \textbf{MT-Bench}
& \textbf{Wild-Bench} \\
\cmidrule(lr){1-1} \cmidrule(lr){2-2} \cmidrule(lr){3-3} \cmidrule(lr){4-4} \cmidrule(lr){5-5}
\textbf{LLaDA-MoE}
& \scoreci{43.7}{-1.6}{+1.5}
& $22.7$
& $2.46$
& $-5.48$ \\
\midrule
\textbf{CE-Only-MoE}
& \scoreci{50.0}{-0.0}{+0.0}
& $22.4$
& $2.84$
& $-5.42$ \\
\textbf{Jitter-MoE}
& \scoreci{48.9}{-1.7}{+1.6}
& $22.2$
& $\mathbf{3.34}$
& $-5.22$ \\
\textbf{CE + CTC-S (MoE)}
& \scoreci{\mathbf{63.7}^{\dagger\ddagger}}{-1.2}{+1.3}
& $\mathbf{25.1}^{\dagger\ddagger}$
& $3.21^{\dagger}$
& $\mathbf{-5.03}^{\dagger\ddagger}$ \\
\bottomrule
\end{tabular}

\label{tab:main_moe}
\end{table}

\section{Comparison to Block Diffusion}
\label{appendix:block-diffusion}
Block diffusion is an inference-time lever that limits the decoding canvas, and a smaller canvas could suppress the occurrence of positional misplacement.
A natural question is whether the gains we observe with CTC-S can be recovered by tuning this lever, i.e., whether CTC-S provides anything beyond what an appropriately tuned block-diffusion baseline already delivers.

To address this, we sweep block sizes $b \in \{32, 64, 128, 256, 512\}$ for LLaDA-8B-Instruct and the CE-Only baseline at inference and report the best per-benchmark score across the sweep. CE~+~CTC-S is run as a single configuration without block-diffusion sweeping. This setup gives the baselines the most favourable inference configuration available under this strategy and asks whether CTC-S still provides additional value on top of it. We restrict the sweep to Creative-Writing-Bench~v3, MT-Bench, and WildBench for computational tractability.\footnote{Arena-Hard requires pairwise judgments against a baseline at every $b$, which inflates the cost.}

We deliberately do not include a CE~+~CTC-S~+~block-diffusion configuration in this comparison. The question this section is designed to answer is whether the training-side lever (CTC-S) provides value beyond what an inference-side lever (block diffusion) can already recover. Answering it cleanly requires varying one factor at a time: a training-side lever (CTC-S, applied to a single inference configuration) against an inference-side lever (block-size sweep applied to models without training-side lever). Combining the two would test a different question, namely whether the training-side and inference-side interventions compose, which we leave to future work.

Table~\ref{tab:block_sweep_ablation} reports the results. Even after sweeping, CE~+~CTC-S is significantly better than the best-swept CE-Only on all three benchmarks and significantly better than the best-swept LLaDA on two of three. The one exception is MT-Bench against LLaDA at $b{=}32$, where the comparison is a statistical tie (mean $\Delta = +0.05$, $\mathrm{CI}_{95}=[-0.29, +0.38]$). Across the six paired comparisons, training-time alignment flexibility outperforms a strictly better-tuned block-diffusion baseline on five and ties on one. This indicates that the contribution of CTC-S is not subsumed by inference-time canvas restriction: relaxing positional supervision during SFT provides gains that reducing the parallelism of decoding does not recover. Full pairwise confidence intervals are reported in Table~\ref{tab:block_sweep_significance}

\begin{table}[t]
\centering
\setlength{\tabcolsep}{4.2pt}
\caption{Block-diffusion sweep ablation. We sweep block sizes $b \in \{32, 64, 128, 256, 512\}$ for LLaDA and CE-Only and report the best score per column (winning $b$ in parentheses); CE + CTC-S is a single non-swept run. $\circ$ and $\diamond$ indicate that CE + CTC-S is significantly better than the best-swept CE-Only and LLaDA variants, respectively (paired bootstrap on aligned sample IDs, 10{,}000 iterations, 95\% CI excludes zero). Bold indicates the best score in each column. The lone non-significant comparison is MT-Bench vs.\ LLaDA at $b{=}32$ (mean $\Delta{=}{+}0.05$, $\mathrm{CI}_{95}{=}[-0.29, +0.38]$).}
\newcommand{\bestblock}[2]{$#1$\,{\scriptsize($b{=}#2$)}}
\begin{tabular}{lccc}
\toprule
 & \textbf{Creative-Writing-}
 & \textbf{MT-}
 & \textbf{Wild-} \\
 & \textbf{Bench v3}
 & \textbf{Bench}
 & \textbf{Bench} \\
\cmidrule(lr){1-1} \cmidrule(lr){2-2} \cmidrule(lr){3-3} \cmidrule(lr){4-4}
\textbf{LLaDA}
 & $23.2$
 & $2.84$
 & $-5.48$ \\
\quad + Block Diffusion (best $b$)
 & \bestblock{26.19}{512}
 & \bestblock{4.30}{32}
 & \bestblock{-4.63}{64} \\
\midrule
\textbf{CE-Only}
 & $24.6$
 & $3.44$
 & $-4.53$ \\
\quad + Block Diffusion (best $b$)
 & \bestblock{26.44}{512}
 & \bestblock{3.53}{512}
 & \bestblock{-4.51}{512} \\
\midrule
\textbf{CE + CTC-S}
 & $\mathbf{27.7}^{\circ\diamond}$
 & $\mathbf{4.35}^{\circ}$
 & $\mathbf{-4.23}^{\circ\diamond}$ \\
\bottomrule
\end{tabular}

\label{tab:block_sweep_ablation}
\end{table}

\section{Robustness to a Shorter Response Length}
\label{app:gen512}
All experiments in \S\ref{sec:experiments} use the fixed inference configuration of \S\ref{subsec:inference} (response length $L=1024$, $N=512$ diffusion steps). To check that the gains of CTC-S are not tied to this canvas, we repeat the open-ended generation comparison on LLaDA-8B-Instruct with $L=512$ and $N=256$ steps, keeping the ratio of steps to response length, the low-confidence remasking strategy, the trained checkpoints, and the evaluation protocol unchanged\footnote{No model is retrained; the three checkpoints of \S\ref{sec:experiments} are decoded with the shorter canvas, and generation and judging otherwise follow \S\ref{sec:experiments}.}.
Table~\ref{tab:gen512} reports the results.
CE~+~CTC-S achieves the best score on three of four benchmarks. It is significantly better than CE-Only ($\dagger$) on Creative-Writing-Bench~v3 and MT-Bench, and significantly better than Jitter ($\ddagger$) on Arena-Hard and MT-Bench.
On Arena-Hard, CE~+~CTC-S does not surpass the CE-Only reference at this length ($-1.1$ pp, 95\% CI $[-4.2, +1.9]$); on WildBench, the three trained models are within each other's confidence intervals.

The pattern observed at $L=1024$ therefore carries over to a shorter canvas: the improvement remains concentrated at the CTC-S step, while the advantage on Arena-Hard is smaller than in Table~\ref{tab:main}.

\begin{table}[t]
\small
\centering
\setlength{\tabcolsep}{4.5pt}
\caption{Performance comparison on open-ended text generation benchmarks on LLaDA-8B-Instruct with response length $L=512$ and $N=256$ diffusion steps. For Arena-Hard, we report bootstrapped 95\% confidence intervals following the official setting. $\dagger$ and $\ddagger$ indicate that CE + CTC-S is significantly better than the CE-Only model and the Jitter model, respectively (paired bootstrap, $p<0.05$). Bold indicates the best score in each column.}
\newcommand{\scoreci}[3]{%
$#1$\ {\scriptsize($#2/#3$)}%
}
\begin{tabular}{lcccc}
\toprule
& \textbf{Arena-Hard\(^*\)}
& \textbf{Creative-Writing-Bench v3}
& \textbf{MT-Bench}
& \textbf{Wild-Bench} \\
\cmidrule(lr){1-1} \cmidrule(lr){2-2} \cmidrule(lr){3-3} \cmidrule(lr){4-4} \cmidrule(lr){5-5}
\textbf{LLaDA}
& \scoreci{44.5}{-1.2}{+1.7}
& $27.5$
& $3.80$
& $-4.64$ \\
\midrule
\textbf{CE-Only}
& \scoreci{\mathbf{50.0}}{-0.0}{+0.0}
& $28.7$
& $3.93$
& $-4.12$ \\
\textbf{Jitter}
& \scoreci{43.8}{-1.3}{+1.4}
& $29.2$
& $4.06$
& $-4.04$ \\
\textbf{CE + CTC-S}
& \scoreci{48.3^{\ddagger}}{-1.7}{+1.5}
& $\mathbf{30.4}^{\dagger}$
& $\mathbf{4.52}^{\dagger\ddagger}$
& $\mathbf{-4.03}$ \\
\bottomrule
\end{tabular}

\label{tab:gen512}
\end{table}

\section{Compute Statement}
\label{appendix:compute-statement}
Most experiments presented in this paper were run on two clusters, both consisting of NVIDIA H100 GPUs with 80~GB of memory. Model training runs on eight H100 GPUs and completes within 10 hours. Generation runs on a single H100 GPU with 80~GB of memory. Evaluation time varies across benchmarks: Arena-Hard takes around 6 hours, MT-Bench typically under 2 hours, Creative-Writing-Bench~v3 around 1.5 hours, and WildBench around 10 hours.

On the same dataset and hardware (eight NVIDIA H100 80GB GPUs), standard CE fine-tuning takes approximately 6.5 hours, while CE + CTC-S fine-tuning takes approximately 10 hours (1.54× wall-clock time). CTC-S also requires more training memory, primarily due to the longer sequences introduced by the additional \textsc{\textless slack\textgreater} tokens rather than the CTC dynamic programming itself. Our method introduces no architectural changes or additional inference-time overhead.

\clearpage  %
\begin{table}[t]
\centering
\caption{Pairwise significance matrix for Table~\ref{tab:main}. Each populated cell reports the 95\% paired-bootstrap confidence interval for the mean score difference (row minus column); $^{*}$ denotes intervals that exclude zero.}
\footnotesize
\setlength{\tabcolsep}{4pt}
\renewcommand{\arraystretch}{1.08}

\begin{tabular*}{\textwidth}{@{\extracolsep{\fill}}lcccc@{}}
\toprule
\multicolumn{5}{@{}l}{\textbf{Arena-Hard}} \\
& \textbf{LLaDA} & \textbf{CE-Only} & \textbf{Jitter} & \textbf{CE+CTC-S} \\
\midrule
\textbf{LLaDA} & -- &  &  &  \\
\textbf{CE-Only} & [0.027, 0.098]$^{*}$ & -- &  &  \\
\textbf{Jitter} & [0.004, 0.091]$^{*}$ & [-0.051, 0.022] & -- &  \\
\textbf{CE+CTC-S} & [0.068, 0.157]$^{*}$ & [0.014, 0.087]$^{*}$ & [0.024, 0.103]$^{*}$ & -- \\
\bottomrule
\end{tabular*}

\vspace{0.6em}

\begin{tabular*}{\textwidth}{@{\extracolsep{\fill}}lcccc@{}}
\toprule
\multicolumn{5}{@{}l}{\textbf{Creative-Writing-Bench v3}} \\
& \textbf{LLaDA} & \textbf{CE-Only} & \textbf{Jitter} & \textbf{CE+CTC-S} \\
\midrule
\textbf{LLaDA} & -- &  &  &  \\
\textbf{CE-Only} & [0.310, 2.434]$^{*}$ & -- &  &  \\
\textbf{Jitter} & [0.763, 3.162]$^{*}$ & [-0.429, 1.585] & -- &  \\
\textbf{CE+CTC-S} & [3.245, 5.869]$^{*}$ & [2.075, 4.316]$^{*}$ & [1.499, 3.735]$^{*}$ & -- \\
\bottomrule
\end{tabular*}

\vspace{0.6em}

\begin{tabular*}{\textwidth}{@{\extracolsep{\fill}}lcccc@{}}
\toprule
\multicolumn{5}{@{}l}{\textbf{MTBench}} \\
& \textbf{LLaDA} & \textbf{CE-Only} & \textbf{Jitter} & \textbf{CE+CTC-S} \\
\midrule
\textbf{LLaDA} & -- &  &  &  \\
\textbf{CE-Only} & [0.125, 1.031]$^{*}$ & -- &  &  \\
\textbf{Jitter} & [0.344, 1.169]$^{*}$ & [-0.212, 0.550] & -- &  \\
\textbf{CE+CTC-S} & [1.075, 1.938]$^{*}$ & [0.594, 1.244]$^{*}$ & [0.412, 1.081]$^{*}$ & -- \\
\bottomrule
\end{tabular*}

\vspace{0.6em}

\begin{tabular*}{\textwidth}{@{\extracolsep{\fill}}lcccc@{}}
\toprule
\multicolumn{5}{@{}l}{\textbf{WildBench}} \\
& \textbf{LLaDA} & \textbf{CE-Only} & \textbf{Jitter} & \textbf{CE+CTC-S} \\
\midrule
\textbf{LLaDA} & -- &  &  &  \\
\textbf{CE-Only} & [0.764, 1.133]$^{*}$ & -- &  &  \\
\textbf{Jitter} & [0.838, 1.201]$^{*}$ & [-0.078, 0.223] & -- &  \\
\textbf{CE+CTC-S} & [1.049, 1.439]$^{*}$ & [0.150, 0.441]$^{*}$ & [0.080, 0.367]$^{*}$ & -- \\
\bottomrule
\end{tabular*}

\label{tab:table1_significance}
\end{table}

\begin{table}[t]
\centering
\caption{Pairwise significance matrix for Table~\ref{tab:main_moe} (LLaDA-MoE). Each populated cell reports the 95\% paired-bootstrap confidence interval for the mean score difference (row minus column); $^{*}$ denotes intervals that exclude zero.}
\footnotesize
\setlength{\tabcolsep}{4pt}
\renewcommand{\arraystretch}{1.08}

\begin{tabular*}{\textwidth}{@{\extracolsep{\fill}}lcccc@{}}
\toprule
\multicolumn{5}{@{}l}{\textbf{Arena-Hard}} \\
& \textbf{LLaDA} & \textbf{CE-Only} & \textbf{Jitter} & \textbf{CE+CTC-S} \\
\midrule
\textbf{LLaDA} & -- &  &  &  \\
\textbf{CE-Only} & [0.027, 0.099]$^{*}$ & -- &  &  \\
\textbf{Jitter} & [0.012, 0.093]$^{*}$ & [-0.046, 0.026] & -- &  \\
\textbf{CE+CTC-S} & [0.160, 0.240]$^{*}$ & [0.102, 0.171]$^{*}$ & [0.107, 0.189]$^{*}$ & -- \\
\bottomrule
\end{tabular*}

\vspace{0.6em}

\begin{tabular*}{\textwidth}{@{\extracolsep{\fill}}lcccc@{}}
\toprule
\multicolumn{5}{@{}l}{\textbf{Creative-Writing-Bench v3}} \\
& \textbf{LLaDA} & \textbf{CE-Only} & \textbf{Jitter} & \textbf{CE+CTC-S} \\
\midrule
\textbf{LLaDA} & -- &  &  &  \\
\textbf{CE-Only} & [-1.462, 0.955] & -- &  &  \\
\textbf{Jitter} & [-1.543, 0.768] & [-1.272, 0.841] & -- &  \\
\textbf{CE+CTC-S} & [1.031, 3.472]$^{*}$ & [1.657, 3.701]$^{*}$ & [1.843, 3.881]$^{*}$ & -- \\
\bottomrule
\end{tabular*}

\vspace{0.6em}

\begin{tabular*}{\textwidth}{@{\extracolsep{\fill}}lcccc@{}}
\toprule
\multicolumn{5}{@{}l}{\textbf{MTBench}} \\
& \textbf{LLaDA} & \textbf{CE-Only} & \textbf{Jitter} & \textbf{CE+CTC-S} \\
\midrule
\textbf{LLaDA} & -- &  &  &  \\
\textbf{CE-Only} & [-0.037, 0.787] & -- &  &  \\
\textbf{Jitter} & [0.481, 1.262]$^{*}$ & [0.181, 0.825]$^{*}$ & -- &  \\
\textbf{CE+CTC-S} & [0.294, 1.188]$^{*}$ & [0.044, 0.688]$^{*}$ & [-0.469, 0.194] & -- \\
\bottomrule
\end{tabular*}

\vspace{0.6em}

\begin{tabular*}{\textwidth}{@{\extracolsep{\fill}}lcccc@{}}
\toprule
\multicolumn{5}{@{}l}{\textbf{WildBench}} \\
& \textbf{LLaDA} & \textbf{CE-Only} & \textbf{Jitter} & \textbf{CE+CTC-S} \\
\midrule
\textbf{LLaDA} & -- &  &  &  \\
\textbf{CE-Only} & [-0.119, 0.234] & -- &  &  \\
\textbf{Jitter} & [0.090, 0.432]$^{*}$ & [0.074, 0.330]$^{*}$ & -- &  \\
\textbf{CE+CTC-S} & [0.260, 0.635]$^{*}$ & [0.252, 0.523]$^{*}$ & [0.064, 0.307]$^{*}$ & -- \\
\bottomrule
\end{tabular*}

\label{tab:main_moe_significance}
\end{table}

\begin{table}[t]
\centering
\caption{Pairwise significance matrix for Table~\ref{tab:block_sweep_ablation} (block-diffusion sweep ablation). Each populated cell reports the 95\% paired-bootstrap confidence interval, on aligned evaluation instances, for the mean score difference between CE+CTC-S (single, non-swept run) and the column baseline at its best block size; $^{*}$ denotes intervals that exclude zero.}
\footnotesize
\setlength{\tabcolsep}{6pt}
\renewcommand{\arraystretch}{1.08}

\begin{tabular*}{\textwidth}{@{\extracolsep{\fill}}lcc@{}}
\toprule
& \textbf{LLaDA + Block Diffusion (best $b$)}
& \textbf{CE-Only + Block Diffusion (best $b$)} \\
\midrule
\textbf{Creative-Writing-Bench v3} & [0.526, 2.535]$^{*}$ & [0.395, 2.237]$^{*}$ \\
\textbf{MT-Bench}                  & [-0.294, 0.381]      & [0.481, 1.169]$^{*}$ \\
\textbf{WildBench}                 & [0.232, 0.559]$^{*}$ & [0.139, 0.422]$^{*}$ \\
\bottomrule
\end{tabular*}

\label{tab:block_sweep_significance}
\end{table}

\clearpage
\begin{table}[t]
\centering
\scriptsize
\tabcolsep 0.4cm
\caption{Existing assets used in this work, with their sources, licenses, and citations. Models, the SFT dataset, and the training framework are listed first, followed by open-ended generation benchmarks (\cref{sec:experiments}) and capability benchmarks (\cref{sec:discussion}).}

\begin{tabular}{ccccc}
\toprule
Asset Type & Asset Name & Link & License & Citation \\
\cmidrule(r){1-1} \cmidrule(r){2-2} \cmidrule(r){3-3} \cmidrule(r){4-4} \cmidrule(r){5-5}
Model & LLaDA-8B-Instruct & \href{https://huggingface.co/GSAI-ML/LLaDA-8B-Instruct}{HuggingFace} & MIT License & \cite{nie2025llada} \\
Model & LLaDA-MoE-7B-A1B-Instruct & \href{https://huggingface.co/inclusionAI/LLaDA-MoE-7B-A1B-Instruct}{HuggingFace} & Apache 2.0 & \cite{zhu2025lladamoe} \\
Dataset & Magpie-Pro-300K-Filtered & \href{https://huggingface.co/datasets/Magpie-Align/Magpie-Pro-300K-Filtered}{HuggingFace} & Llama 3 Community License & \cite{xu2025magpie} \\
Code & d1 (SFT framework) & \href{https://github.com/dllm-reasoning/d1}{Github} & Apache 2.0 & \cite{zhao2025d1} \\
Code & OpenCompass (LLaDA fork) & \href{https://github.com/ML-GSAI/LLaDA/blob/main/EVAL.md}{Github} & Apache 2.0 & \cite{2023opencompass, nie2025llada} \\
Benchmark & Arena-Hard-Auto (v2.0) & \href{https://github.com/lmarena/arena-hard-auto}{Github} & Apache 2.0 & \cite{arenahard2024, li2024crowdsourced} \\
Benchmark & Creative-Writing-Bench v3 & \href{https://github.com/EQ-bench/creative-writing-bench}{Github} & Not specified & \cite{creative-writing-bench-v3} \\
Benchmark & MT-Bench (FastChat) & \href{https://github.com/lm-sys/FastChat/tree/main/fastchat/llm_judge}{Github} & Apache 2.0 & \cite{zheng2023judging} \\
Benchmark & WildBench & \href{https://github.com/allenai/WildBench}{Github} & Apache 2.0 & \cite{lin2025wildbench} \\
Benchmark & GPQA & \href{https://github.com/idavidrein/gpqa}{Github} & MIT License & \cite{rein2024gpqa} \\
Benchmark & MMLU & \href{https://github.com/hendrycks/test}{Github} & MIT License & \cite{hendrycks2021mmlu} \\
Benchmark & MBPP & \href{https://github.com/google-research/google-research/tree/master/mbpp}{Github} & Apache 2.0 & \cite{austin2021mbpp} \\
Benchmark & IFEval & \href{https://github.com/google-research/google-research/tree/master/instruction_following_eval}{Github} & Apache 2.0 & \cite{zhou2023ifeval} \\
Benchmark & GSM8K & \href{https://github.com/openai/grade-school-math}{Github} & MIT License & \cite{cobbe2021gsm8k} \\
Benchmark & MATH500 & \href{https://github.com/openai/prm800k}{Github} & MIT License & \cite{hendrycks2021meath, lightman2024letsverify} \\
\bottomrule
\end{tabular}

\label{tab:assets}
\end{table}

\clearpage
\section*{NeurIPS Paper Checklist}

\begin{enumerate}

\item {\bf Claims}
    \item[] Question: Do the main claims made in the abstract and introduction accurately reflect the paper's contributions and scope?
    \item[] Answer: \answerYes{} %
    \item[] Justification: The abstract and introduction state both contributions delivered in the paper: the controlled shift intervention shows positional sensitivity (\Cref{sec:motivation}) and the CTC-S auxiliary objective achieves consistent gains across four open-ended generation benchmarks (\Cref{sec:experiments}).

\item {\bf Limitations}
    \item[] Question: Does the paper discuss the limitations of the work performed by the authors?
    \item[] Answer: \answerYes{} %
    \item[] Justification: A dedicated Limitations paragraph at the end of \cref{sec:conclusion} discusses the SFT-only scope, the two MDLMs from the LLaDA family, both trained from scratch under the standard masked diffusion objective.

\item {\bf Theory assumptions and proofs}
    \item[] Question: For each theoretical result, does the paper provide the full set of assumptions and a complete (and correct) proof?
    \item[] Answer: \answerNA{} %
    \item[] Justification: The paper does not include formal theoretical results.

    \item {\bf Experimental result reproducibility}
    \item[] Question: Does the paper fully disclose all the information needed to reproduce the main experimental results of the paper to the extent that it affects the main claims and/or conclusions of the paper (regardless of whether the code and data are provided or not)?
    \item[] Answer: \answerYes{} %
    \item[] Justification: \cref{sec:experiments} specifies data, models, benchmarks, and metrics; Appendix~\ref{appendix:imple_details} details the training framework, \textsc{<slack>} insertion, loss computation, and hyperparameters; Appendix~\ref{appendix:ctcs_algo} specifies the CTC-S forward pass.

\item {\bf Open access to data and code}
    \item[] Question: Does the paper provide open access to the data and code, with sufficient instructions to faithfully reproduce the main experimental results, as described in supplemental material?
    \item[] Answer: \answerYes{} %
    \item[] Justification: We attach an anonymized version of the code as supplementary material and will release it publicly upon acceptance.

\item {\bf Experimental setting/details}
    \item[] Question: Does the paper specify all the training and test details (e.g., data splits, hyperparameters, how they were chosen, type of optimizer) necessary to understand the results?
    \item[] Answer: \answerYes{} %
    \item[] Justification: Training and evaluation details (data filtering, model variants, benchmarks, metrics) are reported in \Cref{sec:experiments}; full hyperparameters, optimizer, LoRA configuration, and schedule are in Appendix~\ref{subsec:training_hyperparameters}. We also provide anonymized code as supplementary material.

\item {\bf Experiment statistical significance}
    \item[] Question: Does the paper report error bars suitably and correctly defined or other appropriate information about the statistical significance of the experiments?
    \item[] Answer: \answerYes{} %
    \item[] Justification: Arena-Hard reports the official bootstrapped 95\% confidence intervals; for all benchmarks we additionally report paired-bootstrap 95\% confidence intervals over 10{,}000 iterations on aligned instance-level differences (\Cref{sec:experiments}, Appendix~\ref{appendix:significance}).

\item {\bf Experiments compute resources}
    \item[] Question: For each experiment, does the paper provide sufficient information on the computer resources (type of compute workers, memory, time of execution) needed to reproduce the experiments?
    \item[] Answer: \answerYes{} %
    \item[] Justification: We describe the compute used in Appendix~\ref{appendix:compute-statement}.
    
\item {\bf Code of ethics}
    \item[] Question: Does the research conducted in the paper conform, in every respect, with the NeurIPS Code of Ethics \url{https://neurips.cc/public/EthicsGuidelines}?
    \item[] Answer: \answerYes{} %
    \item[] Justification: The research uses only publicly released artifacts, involves no human subjects, and conforms to the NeurIPS Code of Ethics; see the Ethics statement.

\item {\bf Broader impacts}
    \item[] Question: Does the paper discuss both potential positive societal impacts and negative societal impacts of the work performed?
    \item[] Answer: \answerYes{} %
    \item[] Justification: See the Impact statement.
    
\item {\bf Safeguards}
    \item[] Question: Does the paper describe safeguards that have been put in place for responsible release of data or models that have a high risk for misuse (e.g., pre-trained language models, image generators, or scraped datasets)?
    \item[] Answer: \answerNA{} %
    \item[] Justification: Our modifications to the SFT objective do not introduce capabilities beyond those of the base LLaDA-8B-Instruct model, which is already publicly released.

\item {\bf Licenses for existing assets}
    \item[] Question: Are the creators or original owners of assets (e.g., code, data, models), used in the paper, properly credited and are the license and terms of use explicitly mentioned and properly respected?
    \item[] Answer: \answerYes{} %
    \item[] Justification: All assets are credited and listed in \Cref{tab:assets} with available license information. For Creative-Writing-Bench v3, we were unable to identify a specified license and therefore mark it as ``Not specified''; all other assets are used under their published licenses.

\item {\bf New assets}
    \item[] Question: Are new assets introduced in the paper well documented and is the documentation provided alongside the assets?
    \item[] Answer: \answerNA{} %
    \item[] Justification: The paper does not release new datasets or model checkpoints. Code will be released for reproducibility but is not itself a contributed asset.

\item {\bf Crowdsourcing and research with human subjects}
    \item[] Question: For crowdsourcing experiments and research with human subjects, does the paper include the full text of instructions given to participants and screenshots, if applicable, as well as details about compensation (if any)? 
    \item[] Answer: \answerNA{} %
    \item[] Justification: The paper does not involve crowdsourcing or research with human subjects.

\item {\bf Institutional review board (IRB) approvals or equivalent for research with human subjects}
    \item[] Question: Does the paper describe potential risks incurred by study participants, whether such risks were disclosed to the subjects, and whether Institutional Review Board (IRB) approvals (or an equivalent approval/review based on the requirements of your country or institution) were obtained?
    \item[] Answer: \answerNA{} %
    \item[] Justification: The paper does not involve research with human subjects.

\item {\bf Declaration of LLM usage}
    \item[] Question: Does the paper describe the usage of LLMs if it is an important, original, or non-standard component of the core methods in this research? Note that if the LLM is used only for writing, editing, or formatting purposes and does \emph{not} impact the core methodology, scientific rigor, or originality of the research, declaration is not required.
    \item[] Answer: \answerNA{} %
    \item[] Justification: LLMs are used as judges in the evaluation protocol for open-ended generation benchmarks, which is standard for these benchmarks and not part of the proposed training method. Beyond evaluation, any LLM use was limited to writing and editing.

\end{enumerate}

\end{document}